\documentclass{article}

\usepackage{arxiv}

\usepackage[utf8]{inputenc} 
\usepackage[T1]{fontenc}    
\usepackage{hyperref}       
\usepackage{url}            
\usepackage{booktabs}       
\usepackage{amsfonts}       
\usepackage{nicefrac}       
\usepackage{microtype}      
\usepackage{lipsum}
\usepackage{graphicx}
\usepackage{amsmath} 

\usepackage{cite}
\usepackage{amsmath,amssymb,amsfonts}
\usepackage{algorithmic}
\usepackage{graphicx}
\usepackage{textcomp}
\usepackage{xcolor}
\usepackage{multirow}
\usepackage{amsmath, amsthm}
 \usepackage{colortbl}
 
\graphicspath{ {./images/} }

\title{Bayesian Uncertainty Quantification with Anchored Ensembles for Robust EV Power Consumption Prediction}

\author{
 Ghazal Farhani \\
  \textit{Connected \& Automated Vehicles, National Research Council Canada}\\
  London, Canada\\
  \texttt{ghazal.farhani@nrc-cnrc.gc.ca} \\
   \And
 Taufiq Rahman\\
  \textit{Connected \& Automated Vehicles, National Research Council Canada}\\
  London, Canada\\
  \texttt{Taufiq.Rahman@nrc-cnrc.gc.ca} \\
  \And
 Kieran Humphries \\
 Transportation Emissions \& Electrification Laboratory\\
  Environment and Climate Change Canada\\
  \texttt{Kieran.Humphries@ec.gc.ca} \\
}

\begin{document}
\maketitle
\begin{abstract}
Accurate EV power estimation underpins range prediction and energy management, yet practitioners need \emph{both} point accuracy and trustworthy uncertainty. We propose an anchored-ensemble Long Short-Term Memory (LSTM) with a Student-\(t\) likelihood that jointly captures epistemic (model) and aleatoric (data) uncertainty. Anchoring imposes a Gaussian weight prior (MAP training), yielding posterior-like diversity without test-time sampling, while the \(t\)-head provides heavy-tailed robustness and closed-form prediction intervals. Using vehicle-kinematic time series (e.g., speed, motor RPM), our model attains strong accuracy—RMSE \(3.36\!\pm\!1.10\), MAE \(2.21\!\pm\!0.89\), \(R^2=0.93\!\pm\!0.02\), explained variance \(0.93\!\pm\!0.02\)—and delivers well-calibrated uncertainty bands with near-nominal coverage. Against competitive baselines (Student-\(t\) MC dropout; quantile regression with/without anchoring), our method matches or improves log-scores while producing \emph{sharper} intervals at the same coverage. Crucially for real-time deployment, inference is a single deterministic pass per ensemble member (or a weight-averaged collapse), eliminating Monte Carlo latency. The result is a compact, theoretically grounded estimator that couples accuracy, calibration, and systems efficiency—enabling reliable range estimation and decision-making for production EV energy management.
\end{abstract}


\section{Introduction}

\noindent
\noindent
Road transportation is a major contributor to energy use and emissions globally. In the EU, it accounts for over 70\% of transport-sector greenhouse gases \cite{european2021transport}, while in Canada, transportation represents approximately 25\% of total greenhouse gas emissions, with road transport comprising the majority of this share \cite{eccc2023emissions}. Battery electric vehicles (BEVs) can substantially reduce these emissions through zero tailpipe output, regenerative braking, and improved urban efficiency. However, practical deployment is constrained by driving-range uncertainty, which depends on driver behavior, road grade, 
surface conditions, weather, and battery characteristics (type/age). As transportation electrification accelerates globally, reliable energy management systems with quantified uncertainty become essential for EV adoption and user confidence. Accurate, real-time (or near real-time) 
energy consumption and range estimation are therefore essential for mission planning, eco-routing, and coordinated driving (e.g., platooning) 
\cite{shen2023personalized,madhusudhanan2021computationally,wager2016driving,
zhang2022optimal,zhang2021eco,barhoumi2025fuel}.


\noindent
Energy modeling approaches fall broadly into physics-based and data-driven categories. Physics-based models encode vehicle dynamics from first principles \cite{qu2022urban,fiori2016power,yuan2017method}, while data-driven models learn nonlinear mappings from kinematics to power/energy directly from data \cite{pan2023development,maia2015electrical}. Recent surveys report strong performance of machine learning for EV energy prediction \cite{zhang2024review,chen2021data}, but they also highlight the growing importance of uncertainty quantification (UQ) for deployment: operators require not only point estimates but also calibrated confidence measures. For applications such as range prediction in electric powertrains, especially in mission-critical contexts (e.g., ambulances or police vehicles), operators must not only obtain accurate point predictions but also understand how confident the model is in those predictions. UQ provides this essential capability by characterizing two key types of uncertainty: \emph{aleatoric}, arising from inherent variability in driving conditions, sensor noise, and exogenous factors, and \emph{epistemic}, stemming from limited, biased, or non-representative training data \cite{valdenegro2022deeper}.

\noindent
In prior work \cite{yahyaabadi2025deep}, we benchmarked Long Short-Term Memory (LSTM) networks against Temporal Convolutional Networks (TCNs), Transformers, and Random Forest on EV telematics for power prediction and found that LSTMs achieved comparable accuracy (RMSE/MAE) to the more complex architectures. To avoid re-opening architecture comparisons in this paper, we focus exclusively on LSTMs and refer readers to \cite{yahyaabadi2025deep} for the full benchmarking protocol and results. This focus also facilitates our Bayesian treatment of weights and makes weight evolution and calibration diagnostics more interpretable within a recurrent setting.

\noindent
In this paper by using real-world telematics, we estimate instantaneous power consumption with an LSTM and perform a comprehensive UQ analysis. We adopt a Bayesian ensembling method \cite{pearce2020uncertainty} to capture \emph{epistemic} uncertainty and extend its formulation to LSTM architectures; we additionally incorporate a probabilistic output layer to model \emph{aleatoric} uncertainty, yielding predictive intervals suitable for online range management. The resulting uncertainty bands track error scales and provide actionable confidence bounds for battery management and eco-driving.

\noindent
\textbf{Contributions:} (i) We develop an LSTM-based EV power estimator with principled uncertainty quantification (UQ) that yields calibrated prediction intervals. (ii) We provide a mathematical extension of a Bayesian ensembling scheme—\emph{anchored networks}, originally proposed for feedforward models—to recurrent LSTM architectures, and derive the corresponding training objective. (iii) We validate on real telematics data and report both accuracy and calibration metrics. (iv) To isolate model-uncertainty effects, we benchmark epistemic UQ via anchored networks against MC dropout \cite{gal2016dropout}. (v) For aleatoric UQ, we demonstrate that a Student’s-$t$ negative log-likelihood loss provides superior calibration and accuracy compared to quantile regression.

\noindent
\textbf{Paper Organization:}
Section~\ref{sec:related_work} reviews learning-based BEV power estimation and UQ, and identifies gaps in the literature. Section~\ref{sec:Bayesian_methodology} presents the Bayesian anchored-ensemble methodology, sketches its extension to LSTM networks, introduces the aleatoric loss, and derives the complete training objective. Section~\ref{sec:Data-Collection} briefly describes the dataset; since part of it is detailed elsewhere \cite{yahyaabadi2025deep}, we focus here on the chassis-dynamometer experiments, feature selection, and network architecture. Section~\ref{sec:result} reports results and compares (i) anchored ensembles with a Student’s-$t$ negative log-likelihood, (ii) anchored ensembles with quantile loss, (iii) MC dropout with a Student’s-$t$ negative log-likelihood, and (iv) MC dropout with quantile loss. We show that the Student’s-$t$ loss yields better-calibrated intervals than quantile loss, and that anchored ensembles achieve performance comparable to MC dropout while offering a more practical choice for our setting. Section~\ref{sec:conclusion} concludes.

\section{Related Work}\label{sec:related_work}
This section reviews deep learning methods for battery electric vehicle (BEV) power/energy estimation and summarizes prior efforts on uncertainty quantification (UQ) in this context.

Early data-driven studies employed feedforward neural networks (FNNs) for segment-level energy prediction. For example, \cite{de2017data} combined multiple linear regression with a neural network using kinematic and road-context features (e.g., speed, acceleration, road condition), reporting mean absolute error (MAE) in the 12--14\% range. Recurrent architectures, notably Long Short-Term Memory (LSTM) networks, have since shown stronger performance by exploiting temporal dependencies. Chen et al. \cite{chen2021data} trained an LSTM on instantaneous speed, acceleration, and road grade, achieving approximately 3\% mean absolute percentage error (MAPE) for cumulative energy use.

Uncertainty quantification (UQ) is essential across many scientific and engineering disciplines, as it provides confidence levels for model predictions. From optimal inverse estimation methods to neural networks, the importance of UQ in numerical modeling has been well established \cite{farhani2019optimal, farhani2023bayesian, sullivan2015introduction, abdar2021review}. Nevertheless, UQ applications in battery electric vehicle (BEV) energy modeling remain sparse. One prevalent approach substitutes mean-squared error with quantile loss to generate prediction intervals. For instance, Chen et al. \cite{chen2023probabilistic} employed Quantile LSTM (QLSTM) to achieve 80\% and 90\% coverage for BEV charging-demand forecasts, while Zhu et al. \cite{zhu2024predicting} applied quantile recurrent neural networks (QRNN) to predict cumulative energy at the 90\% confidence level. Quantile regression has similarly proven popular in related energy forecasting tasks, such as wind power prediction \cite{faustine2022fpseq2q}. Despite being model-agnostic and computationally tractable, quantile-based methods predominantly capture aleatoric uncertainty (irreducible data noise) while neglecting epistemic uncertainty (reducible model uncertainty arising from insufficient or unrepresentative training data). Furthermore, these approaches do not produce complete predictive distributions, lack natural interpolation between specified quantiles, and require computational effort that scales linearly with the number of quantiles estimated.

To capture epistemic uncertainty, Monte Carlo (MC) dropout has been widely adopted (e.g., \cite{yalamanchi2023uncertainty,hashemipour2022uncertainty,wang2020survey}), following the insight of \cite{gal2016dropout} that applying dropout at inference approximates Bayesian model uncertainty.

Beyond BEV-specific literature, a substantial body of work addresses UQ more broadly in machine learning. Bayesian theory (e.g., \textit{MacKay, 1992} \cite{mackay1992practical})  provides a principled framework by defining posterior distributions over model parameters, thereby embedding epistemic uncertainty directly in the model and enabling the use of informative priors. In practice, however, full Bayesian inference for modern deep networks is computationally prohibitive; Markov chain Monte Carlo (MCMC) \cite{van2018simple}, while accurate, is rarely tractable at scale.

Ensemble-based methods offer a practical alternative but are often heuristic. The Bayesian ensembling approach of \cite{pearce2020uncertainty} provides a principled alternative grounded in Bayesian point estimation and has shown practical utility across domains \cite{farhani2023bayesian}. Originally developed for feedforward networks, we present a natural extension of its mathematical formulation to LSTM architectures in this work.

\section{Methodology: Bayesian Anchored Ensembles for LSTMs} \label{sec:Bayesian_methodology}
\label{sec:material_and_methods}
In this section, we briefly describe the Bayesian anchored ensemble method, then state a proposition showing that the anchoring idea extends naturally to LSTM models. Finally, we present a unified loss that jointly estimates aleatoric and epistemic uncertainties within the power-estimation pipeline.

\subsection{Bayesian Anchored Ensemble for Epistemic Uncertainty}
We model epistemic uncertainty via a weight prior and MAP training. With parameters $W$ and data $\mathcal{D}=\{(x_n,y_n)\}_{n=1}^N$, the average negative log-posterior is
\begin{equation}
\label{eq:map-obj}
\mathcal{J}(W)
=\underbrace{\frac{1}{N}\sum_{n=1}^N \ell\!\big(\hat y(x_n;W),y_n\big)}_{\mathcal{L}_{\mathrm{data}}(W)}
+\underbrace{\tfrac{1}{2}(W-\mu_{\!p})^\top \Sigma_{\!p}^{-1}(W-\mu_{\!p})}_{\mathcal{L}_{\mathrm{prior}}(W)} ,
\end{equation}
where $\ell$ is the per-sample NLL (e.g., Student-$t$), and $p(W)=\mathcal{N}(\mu_{\!p},\Sigma_{\!p})$.
Anchored ensembles \cite{pearce2020uncertainty} draw anchors $W_{\text{anc}}^{(m)}\!\sim\!\mathcal{N}(\mu_{\!p},\Sigma_{\!p})$ and train members by replacing $\mu_{\!p}$ with $W_{\text{anc}}^{(m)}$ in $\mathcal{L}_{\mathrm{prior}}$; this yields posterior-like diversity without test-time sampling.

\subsection{Extension to LSTMs}
\label{sec:Extension_LSTM}
Let the LSTM parameters be partitioned by gates/layers as $W=\bigoplus_{g\in\mathcal{G}} W_g$, $\mathcal{G}=\{i,f,o,c,\text{head}\}$. Using a factorized Gaussian prior
$p(W)=\prod_{g\in\mathcal{G}}\mathcal{N}\!\big(W_g;W_g^{(0)},\sigma_g^2 I\big)$, the MAP objective \eqref{eq:map-obj} becomes
\begin{equation}
\label{eq:lstm-map}
\mathcal{J}(W)=\mathcal{L}_{\mathrm{data}}(W)+\sum_{g\in\mathcal{G}}\frac{1}{2\sigma_g^2}\,\big\|W_g-W_g^{(0)}\big\|_2^2,
\end{equation}
i.e., the data term depends on \emph{all} gates via the recurrence, while the prior contributes independent quadratic penalties per gate. In anchored ensembles, draw gate-wise anchors $W_{g,\text{anc}}^{(m)}$ and replace $W_g^{(0)}$ by $W_{g,\text{anc}}^{(m)}$ in \eqref{eq:lstm-map} for each member $m$.

\paragraph*{Proposition (MAP with gate-wise Gaussian priors)}
Under the sample-factorized likelihood and block-factorized prior above, any MAP estimator minimizes \eqref{eq:lstm-map}. \emph{Proof:} See Appendix~\ref{sec:appendic}.

\subsection{Aleatoric Uncertainty: Heteroscedastic Gaussian Negative Log-Likelihood}

To learn aleatoric (data-dependent) noise, it is common to minimize the heteroscedastic Gaussian NLL \cite{gal2016dropout}, however, when residuals are heavy-tailed, a Student’s $t$ likelihood is more robust:
\begin{align}
\mathcal{L}_{t}
= \frac{1}{N}\sum_{i=1}^{N}\Big[
\log s(\mathbf{x}_i) + \tfrac{1}{2}\log(\nu\pi)
+ \log\Gamma(\tfrac{\nu}{2}) - \log\Gamma(\tfrac{\nu+1}{2}) \nonumber\\[-2pt]
\qquad\qquad
+ \tfrac{\nu+1}{2}\log\!\Big(1 + \frac{(y_i-\hat y(\mathbf{x}_i))^2}{\nu\,s(\mathbf{x}_i)^2}\Big)
\Big],
\label{eq:student_t_nll}
\end{align}
where $s(\mathbf{x})>0$ is the predicted scale and $\nu>0$ the degrees of freedom. As $\nu\!\to\!\infty$, \eqref{eq:student_t_nll} approaches the Gaussian distribution; if $\nu>2$, $\mathrm{Var}[Y\mid\mathbf{x}]=\frac{\nu}{\nu-2}\,s(\mathbf{x})^2$.

\subsection{Aleatoric Uncertainty and Combined Objective}
For heavy-tailed residuals we use a Student-$t$ head with location $\mu(x)$, scale $s(x)\!>\!0$, and dof $\nu$. The per-sample $t$-NLL (omitting constants) is
$\ell_t\big(y\mid \mu,s,\nu\big)=\log s+\tfrac{\nu+1}{2}\log\!\big(1+\tfrac{(y-\mu)^2}{\nu\,s^2}\big)$.
We train an anchored ensemble with the unified \(t\)+prior objective (member $m$):
\begin{equation}
\label{eq:combined-t-map}
\mathcal{J}^{(m)}=\tfrac1N\sum_{n}\ell_t\!\big(y_n\mid \mu^{(m)}_n,s^{(m)}_n,\nu\big)
+\sum_{g\in\mathcal{G}}\tfrac{\|W^{(m)}_g-W^{(m)}_{g,\text{anc}}\|_2^2}{2\sigma_g^2}.
\end{equation}

Predictive intervals at level $1-\alpha$ use $t_{1-\alpha/2,\nu}\,s^{(m)}(x)$ and ensemble averaging for epistemic spread.

\subsection{Evaluation metrics}
The performance of the time-series regression model is evaluated using two standard metrics: Mean Absolute Error (MAE) and Root Mean Square Error (RMSE). The MAE is defined as
\begin{equation}
\text{MAE} = \frac{1}{N} \sum_{i=1}^{N} \left| y_i - \hat{y}_i \right|,
\end{equation}
while the RMSE is given by
\begin{equation}
\text{RMSE} = \sqrt{\frac{1}{N} \sum_{i=1}^{N} (y_i - \hat{y}_i)^2}.
\end{equation}
In both expressions, \( y_i \) denotes the ground truth and \( \hat{y}_i \) the predicted value from the neural network model for the \( i \)-th sample, and \( N \) is the total number of samples.

\section{Data Collection and Model Architecture}
This section outlines the experimental data collection, details the model architecture selection, and concludes with the feature selection approach.

\subsection{Data Collection} \label{sec:Data-Collection}

To probe model performance under both idealized and realistic operating conditions, we assembled two complementary datasets. The first comprises four controlled chassis–dynamometer tests following the Highway Fuel Economy Test (HWFET) cycle—two sedans, one hatchback, and one truck. Each HWFET run lasts 800\,s and covers 16.45\,km at an average speed of 77.7\,km/h. The laboratory setting minimizes exogenous variability (e.g., road friction changes, wind, temperature gradients), providing a high–fidelity environment to learn the mapping between vehicle kinematics and power consumption. From each vehicle we extracted signals such as velocity, acceleration, resistive force, driving (tractive) force, and related measures. Although all four vehicles share the same velocity trajectory by design, their instantaneous power profiles differ; consequently, for each vehicle we trained on 70\% of the samples and evaluated on the remaining 30\%.

To assess external validity and examine how predictive uncertainty evolves in realistic highway driving, we also collected a real–world dataset from a 2022 Hyundai IONIQ 5 on Ontario Highway~403 (London $\to$ Sarnia). Vehicle telematics were captured using the ``Car Scanner'' smartphone application through the OBD-II port, recording parameters reported by the vehicle. Unlike the dyno data, these measurements—and the acquisition method—are inherently noisier and subject to uncontrolled factors, offering a more representative testbed of on–road conditions and commodity sensing. This pairing of datasets—highly controlled vs.\ real–world—allows us to study both the learnability of the model in ideal conditions and the growth (or drift) of estimation uncertainty under practical deployment.

\subsection{Model Architecture Selection}
The model architecture and training hyperparameters were selected through randomized trials and empirical evaluation. Specifically, the number of LSTM layers was randomly chosen from \{2, 4, 6, 8\}, while the hidden dimension was selected from \{32, 64, 128\}. The anchored variance for each LSTM gate was sampled from a uniform range of $[0.001, 0.1]$. The number of training epochs was selected from \{50, 100, 300, 500\}, and the learning rate was randomly sampled from the range $[10^{-5}, 10^{-2}]$. Optimization algorithms tested included Adam, stochastic gradient descent, AdaGrad, and RMSprop.

To balance model performance with computational efficiency, the final configuration employed a four-layer LSTM architecture with a hidden dimension of 32 and an anchored variance of 0.01. Since all input features were normalized to the $[0, 1]$ range, this choice of anchored variance was considered appropriate, ensuring that the learned weights remained on a comparable scale. For ensemble training, 30 models with identical architecture were trained independently, each for 300 epochs using a learning rate of 0.001 and the Adam optimization algorithm.

\subsection{Feature Selection}
Following the feature--selection process described in \cite{yahyaabadi2025deep}, we adopted the same set of features since the highway dataset used in our work is identical to that study. While some prior works (e.g., \cite{nabi2023parametric, achariyaviriya2023estimating}) included battery current as an input, we followed the approach of \cite{yahyaabadi2025deep} and relied solely on vehicle kinematics. This design choice ensures that the deep learning model learns the intrinsic relationship between vehicle telemetry and battery power consumption without depending on electrical measurements. Accordingly, the core features are vehicle velocity, acceleration, and motor torque.  

In the chassis--dynamometer experiments, multiple signals were recorded; however, consistent with our decision to exclude electrical variables (\texttt{current}/\texttt{voltage}), the candidate input set was limited to: longitudinal speed (\texttt{Speed}), smoothed longitudinal acceleration (\texttt{Acceleration}), tractive (drive) force at the wheels/axle (\texttt{DY\_flt\_force}), road--load resistance (aerodynamic, rolling, and grade) (\texttt{DY\_Roadld}), parasitic/drivetrain and accessory losses (\texttt{DY\_Parasitic}), and driver/ECU drive request (\texttt{DY\_DriveRef}). Randomized feature--subset evaluations were conducted on more than 15 different configurations; representative results are summarized in Table~\ref{tab:feature}.  

As shown in the table, the feature combination in the first row achieved the best performance, with RMSE = 2.4 and MAE = 1.7. Acceleration emerged as a critical predictor: excluding it caused RMSE and MAE to rise sharply (8.1 and 6.6, respectively). To further test this, we trained a model using only acceleration as input (Row 5). Although acceleration alone contributed substantially, the error metrics increased markedly, confirming that the additional features in Row~1 collectively yield the lowest RMSE and MAE.

\begin{table}[t]
\centering
\resizebox{0.3\linewidth}{!}{%
\begin{tabular}{|ll|l|l|}
\hline
\rowcolor[HTML]{ECF4FF}
\multicolumn{2}{|l|}{\cellcolor[HTML]{ECF4FF}Feature Set} & RMSE & MAE \\ \hline
\multicolumn{1}{|l|}{1} & Set 1 & \textbf{2.4} & \textbf{1.7} \\ \hline
\multicolumn{1}{|l|}{2} & Set 2 & 2.5 & \textbf{1.7} \\ \hline
\multicolumn{1}{|l|}{3} & Set 3 & 8.1 & 6.6 \\ \hline
\multicolumn{1}{|l|}{4} & Set 4 & 11.2 & 9.2 \\ \hline
\multicolumn{1}{|l|}{5} & Set 5 & 5.5 & 3.3 \\ \hline
\end{tabular}%
}
\caption{Ablation on feature subsets. 
S1: \{DY\_DriveRef, DY\_Parasitic, DY\_flt\_force, Speed, Acceleration\};
S2: \{DY\_DriveRef, DY\_Parasitic, DY\_Roadld, DY\_flt\_force, Speed, Acceleration\};
S3: \{DY\_DriveRef, DY\_Parasitic, DY\_Roadld, DY\_flt\_force, Speed\};
S4: \{DY\_DriveRef, DY\_flt\_force\};
S5: \{Acceleration\}.}
\label{tab:feature}
\end{table}

\section{Result} \label{sec:result}
\subsection{The Anchored LSTM Model}
\begin{figure}
    \centering
    \includegraphics[width= 0.7\linewidth]{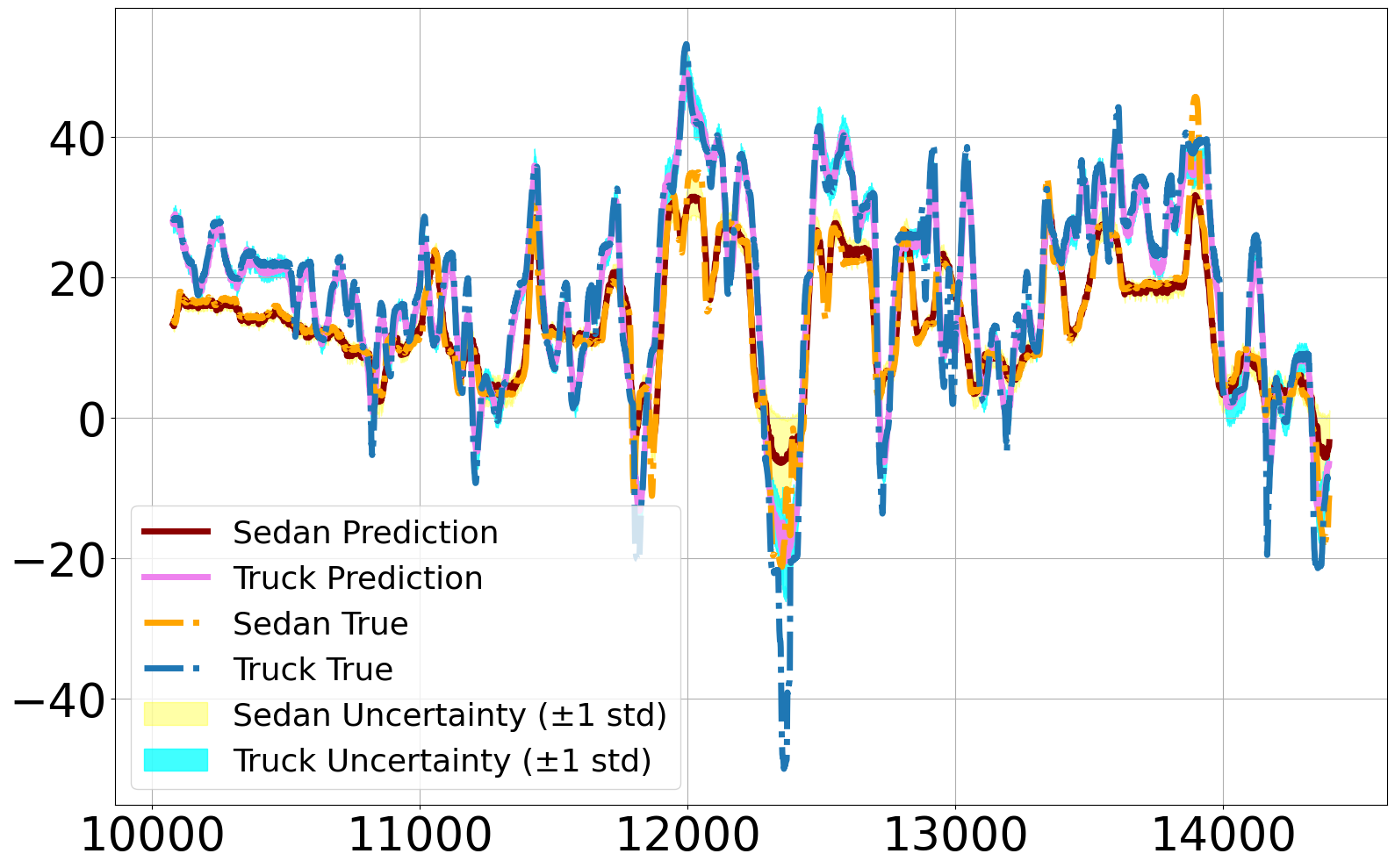}
    \caption{Estimated power consumption for the sedan (solid dark red) and the truck (solid violet); ground truth shown as dashed orange and dashed blue, respectively. Confidence intervals are depicted by the shaded yellow and green bands.}

    \label{fig:result-chassis}
\end{figure}

The anchored LSTM was implemented in Python with Keras and trained on an NVIDIA T4 GPU. After training, we evaluated the model on a held-out test set. Table II reports RMSE, MAE, $R^2$, and explained variance for four vehicles using the chassis dynamometer experiments and the IONIQ 5 measurements. For visualization, Fig.~\ref{fig:result-chassis} shows two representative cases—a sedan and a truck. Dashed lines denote ground truth, solid lines the anchored model’s predictions, and the shaded bands the combined (aleatoric + epistemic) uncertainty intervals. Although the chassis dynamometer velocity profiles are identical, differences across fleets lead to distinct power demands; accordingly, each model is trained for a fleet-specific pattern rather than a single shared mapping.

\subsection{Comparative Study}
Given the widespread use of quantile regression for aleatoric uncertainty \cite{chen2021data,de2017data} and MC dropout for epistemic uncertainty \cite{gal2016dropout}, we include three comparison baselines: (i) \emph{Quantile–Anchor}: quantile loss with our anchored ensemble; (ii) \emph{Quantile–Dropout}: quantile loss with MC dropout; and (iii) \emph{t–NLL Dropout}: Student’s-$t$ negative log-likelihood with MC dropout. These variants allow a systematic assessment of aleatoric modeling (quantiles vs.\ $t$-likelihood) and epistemic modeling (anchoring vs.\ dropout), alongside our primary \emph{t–NLL Anchor} model.

\begin{figure}
    \centering
    \includegraphics[width=0.45\linewidth]{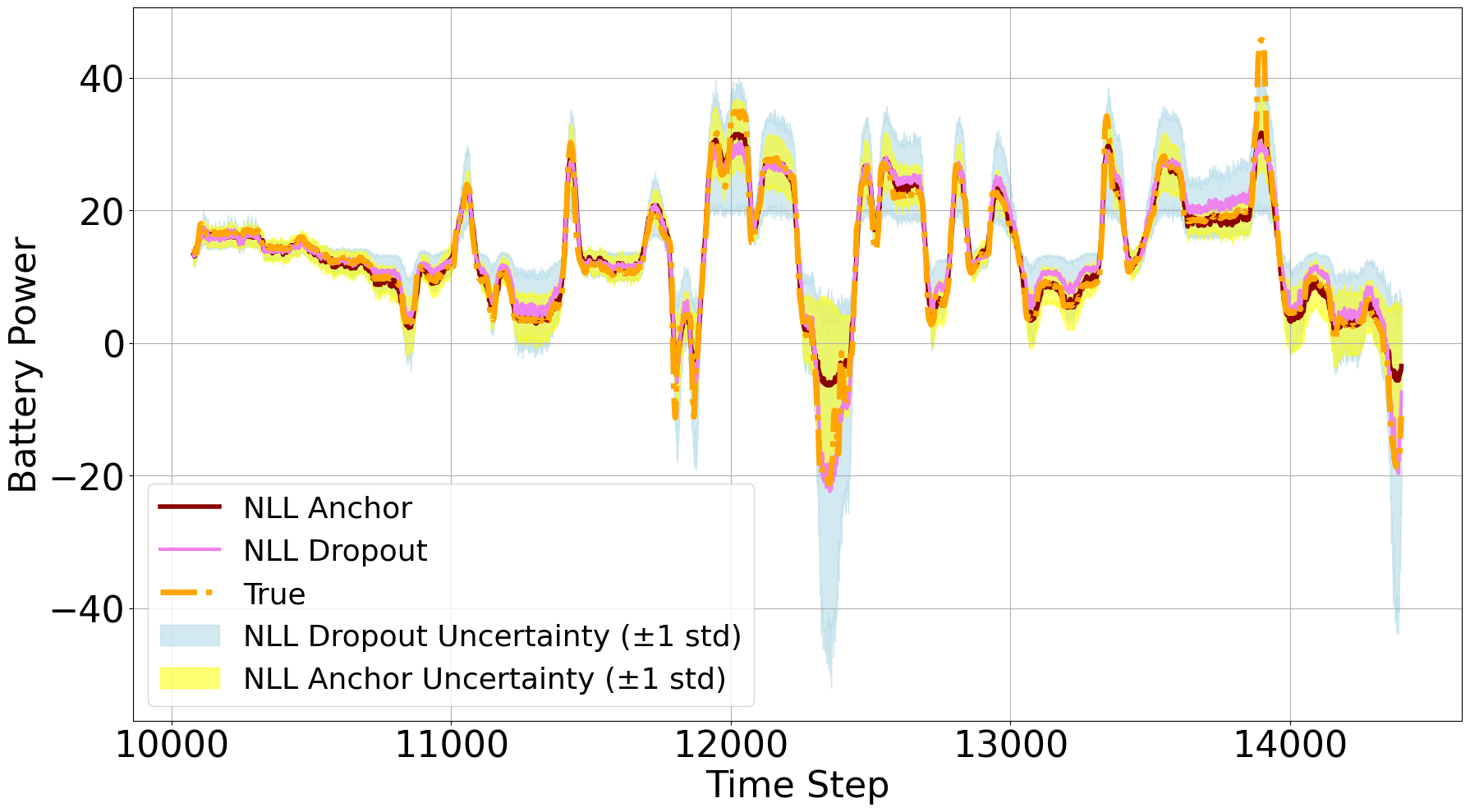}
     \includegraphics[width=0.45\linewidth]{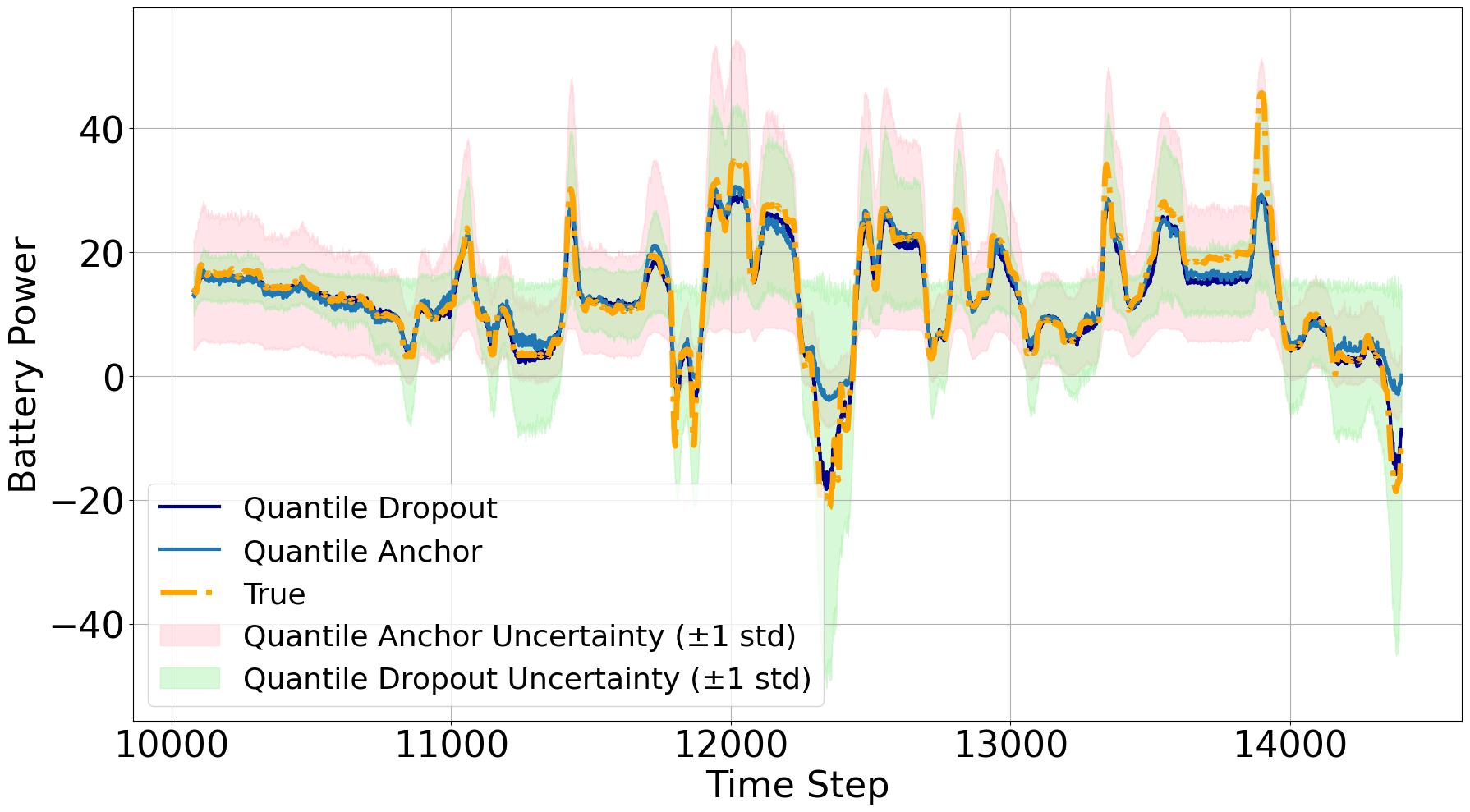}
    \caption{Left Panel: \emph{t–NLL Anchor} (dark red) and \emph{t–NLL Dropout} (violet), with their confidence intervals (yellow and light blue bands)  Right Panel:  \emph{Quantile–Anchor} (dark blue) and \emph{Quantile–Dropout} (light blue) with corresponding bands (light pink and light green)}
    \label{fig:uncertainty_Sedan}
\end{figure}

We report RMSE, MAE, $R^2$, and explained variance for four vehicles using chassis-dynamometer experiments and highway tests by ION5 in Table II. For visualization, Fig.~\ref{fig:uncertainty_Sedan} shows Sedan results. The left panel compares \emph{t–NLL Anchor} (dark red) and \emph{t–NLL Dropout} (violet), with their confidence intervals (yellow and light blue bands). Consistent with Table II, both achieve low MAE/RMSE and high $R^2$ and explained variance, with similar aleatoric and epistemic uncertainty. The right panel presents \emph{Quantile–Anchor} (dark blue) and \emph{Quantile–Dropout} (light blue) with corresponding bands (light pink and light green); although errors remain low, intervals are noticeably wider. 

To evaluate calibration, we apply the binomial proportion confidence-interval test \cite{vollset1993confidence} at $\alpha=0.1$ (90\% target). Table~\ref{tab:confidence} reports empirical coverage with its CI bounds, average band width, and a standardized variance metric (ideal $=1$; $>1$ indicates under-dispersion/narrow bands; $<1$ over-dispersion/wide bands). Moreover, \emph{t–NLL Anchor} attains coverage within the bounds with standardized variance near 1 (well calibrated); \emph{t–NLL Dropout} is slightly conservative (standardized variance $<1$ and coverage near the upper bound); both quantile variants are overly conservative (standardized variance $\ll 1$ with substantially wider bands).
   
\begin{figure}
    \centering
    \includegraphics[width=0.45\linewidth]{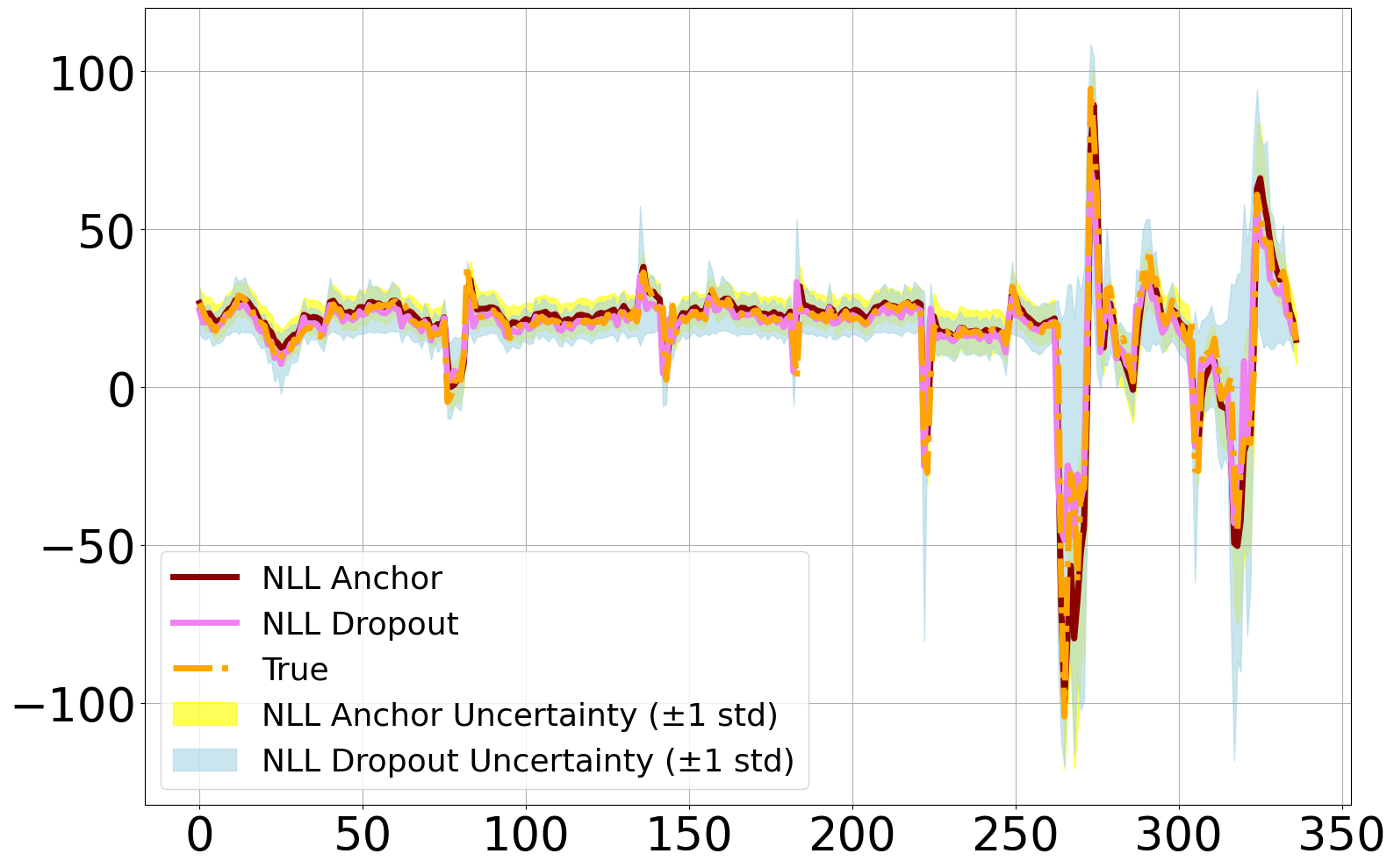}
    \includegraphics[width=0.45\linewidth]{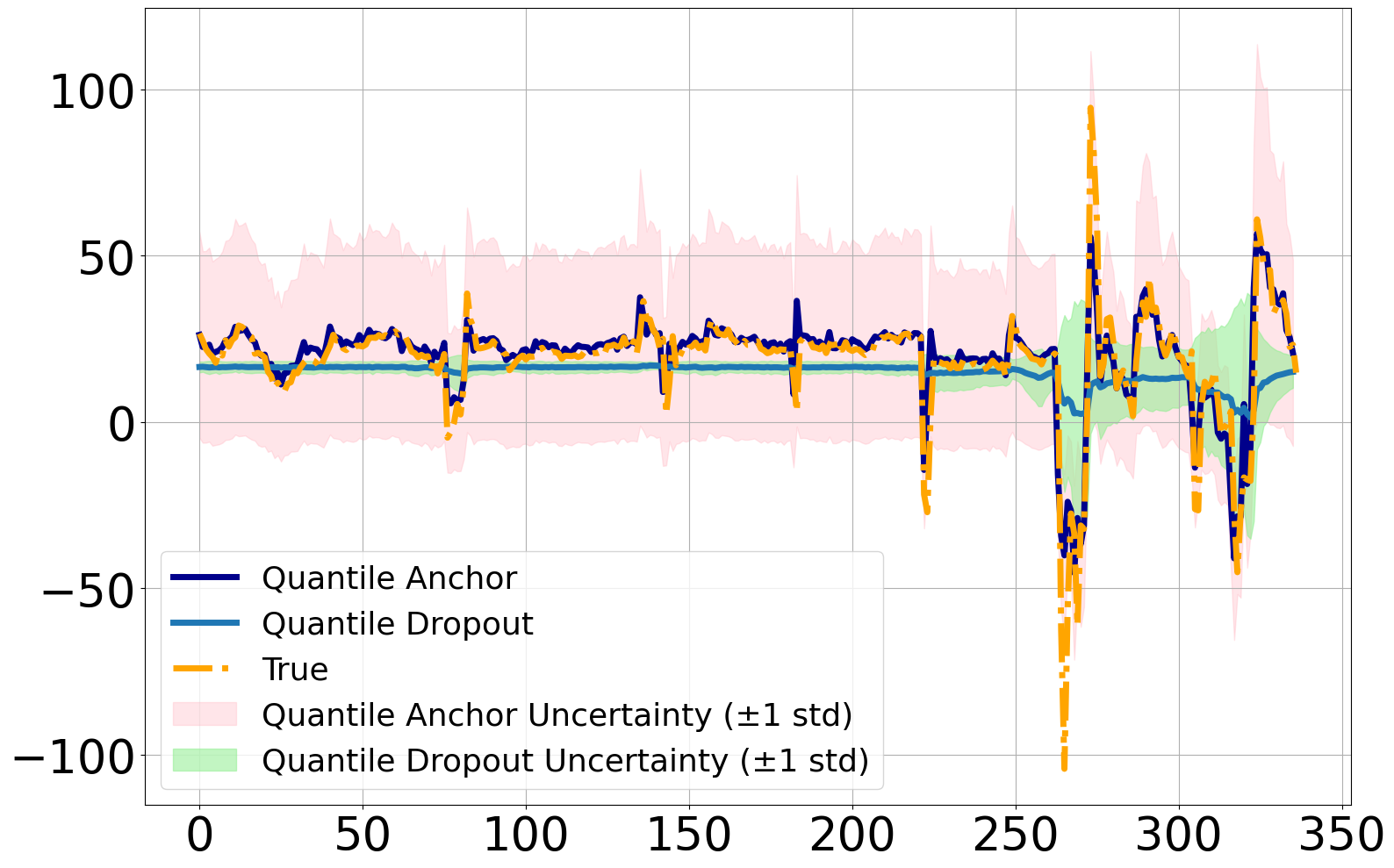}
    \caption{Left Panel: \emph{t–NLL Anchor} (dark red) and \emph{t–NLL Dropout} (violet), with their confidence intervals (yellow and light blue bands)  Right Panel:  \emph{Quantile–Anchor} (dark blue) and \emph{Quantile–Dropout} (light blue) with corresponding bands (light pink and light green)}
\label{fig:_Ionic_all}
\end{figure}

Figure~\ref{fig:_Ionic_all} summarizes the IONIQ 5 highway experiments. The left panel compares ground truth with the t–NLL Anchor and t–NLL Dropout predictions (with their confidence bands), while the right panel shows the Quantile–Anchor and Quantile–Dropout counterparts. Because these data were collected on open highways via a mobile app (as opposed to controlled chassis–dynamometer tests; see \cite{yahyaabadi2025deep}), RMSE/MAE are slightly higher overall. Even so, the t–NLL models maintain low errors and good calibration (Table~\ref{tab:confidence}). In contrast, Quantile–Dropout exhibits noticeably larger errors and misses fine-scale structure, whereas Quantile–Anchor fits more details but produces overly wide intervals (e.g., $\sim$95\% bands with standardized variance $\approx 0.10$), indicating under-confidence. Calibration metrics further favor anchoring: t–NLL Anchor attains standardized variance near unity ($\approx 1.13$), while t–NLL Dropout is more under-dispersed ($\approx 1.45$; narrower-than-ideal bands), making the anchored approach the more suitable choice in this setting.

\begin{table*}[]
\centering
\begin{tabular}{|
>{\columncolor[HTML]{FFEAE9}}c c c c c c c|
>{\columncolor[HTML]{FFEAE9}}c c c c c c c|}
\hline
\multicolumn{7}{|c|}{\cellcolor[HTML]{9AFF99}NLL Anchor} &
\multicolumn{7}{c|}{\cellcolor[HTML]{9AFF99}Quantile Drop Out} \\ \hline

\multicolumn{1}{|c|}{\cellcolor[HTML]{CBCEFB}Vehicle} &
\multicolumn{1}{c|}{\cellcolor[HTML]{DCFFDC}RMSE} &
\multicolumn{1}{c|}{\cellcolor[HTML]{DCFFDC}MAE} &
\multicolumn{1}{c|}{\cellcolor[HTML]{DCFFDC}AU}  &
\multicolumn{1}{c|}{\cellcolor[HTML]{DCFFDC}EU}  &
\multicolumn{1}{c|}{\cellcolor[HTML]{DCFFDC}$R^2$}  &
\cellcolor[HTML]{DCFFDC}EV &
\multicolumn{1}{c|}{\cellcolor[HTML]{CBCEFB}Vehicle} &
\multicolumn{1}{c|}{\cellcolor[HTML]{DCFFDC}RMSE} &
\multicolumn{1}{c|}{\cellcolor[HTML]{DCFFDC}MAE} &
\multicolumn{1}{c|}{\cellcolor[HTML]{DCFFDC}AU}  &
\multicolumn{1}{c|}{\cellcolor[HTML]{DCFFDC}EU}  &
\multicolumn{1}{c|}{\cellcolor[HTML]{DCFFDC}$R^2$}  &
\cellcolor[HTML]{DCFFDC}EV \\ \hline

\multicolumn{1}{|c|}{\cellcolor[HTML]{FFEAE9}Sedan\_1}  &
\multicolumn{1}{c|}{2.5} & \multicolumn{1}{c|}{1.3} & \multicolumn{1}{c|}{1.6} &
\multicolumn{1}{c|}{1.2} & \multicolumn{1}{c|}{0.93} & 0.94 &
\multicolumn{1}{c|}{\cellcolor[HTML]{FFEAE9}Sedan\_1}  &
\multicolumn{1}{c|}{2.5} & \multicolumn{1}{c|}{1.6} & \multicolumn{1}{c|}{5.5} &
\multicolumn{1}{c|}{6.5} & \multicolumn{1}{c|}{0.93} & 0.94 \\ \hline

\multicolumn{1}{|c|}{\cellcolor[HTML]{FFEAE9}Sedan\_2}  &
\multicolumn{1}{c|}{2.3} & \multicolumn{1}{c|}{1.7} & \multicolumn{1}{c|}{1.3} &
\multicolumn{1}{c|}{1.35} & \multicolumn{1}{c|}{0.90} & 0.94 &
\multicolumn{1}{c|}{\cellcolor[HTML]{FFEAE9}Sedan\_2}  &
\multicolumn{1}{c|}{2.7} & \multicolumn{1}{c|}{1.7} & \multicolumn{1}{c|}{6.5} &
\multicolumn{1}{c|}{7.8} & \multicolumn{1}{c|}{0.89} & 0.93 \\ \hline

\multicolumn{1}{|c|}{\cellcolor[HTML]{FFEAE9}Truck}     &
\multicolumn{1}{c|}{3.5} & \multicolumn{1}{c|}{2.0} & \multicolumn{1}{c|}{1.5} &
\multicolumn{1}{c|}{1.4} & \multicolumn{1}{c|}{0.95} & 0.95 &
\multicolumn{1}{c|}{\cellcolor[HTML]{FFEAE9}Truck}     &
\multicolumn{1}{c|}{5.7} & \multicolumn{1}{c|}{4.3} & \multicolumn{1}{c|}{6.9} &
\multicolumn{1}{c|}{7.9} & \multicolumn{1}{c|}{0.87} & 0.87 \\ \hline

\multicolumn{1}{|c|}{\cellcolor[HTML]{FFEAE9}Hatchback} &
\multicolumn{1}{c|}{3.1} & \multicolumn{1}{c|}{2.1} & \multicolumn{1}{c|}{0.95} &
\multicolumn{1}{c|}{1.45} & \multicolumn{1}{c|}{0.89} & 0.88 &
\multicolumn{1}{c|}{\cellcolor[HTML]{FFEAE9}Hatchback} &
\multicolumn{1}{c|}{4.5} & \multicolumn{1}{c|}{3.4} & \multicolumn{1}{c|}{3.7} &
\multicolumn{1}{c|}{4.7} & \multicolumn{1}{c|}{0.76} & 0.78 \\ \hline

\multicolumn{1}{|c|}{\cellcolor[HTML]{FFEAE9}IONIQ 5}   &
\multicolumn{1}{c|}{5.4} & \multicolumn{1}{c|}{3.9} & \multicolumn{1}{c|}{7.0} &
\multicolumn{1}{c|}{3.5} & \multicolumn{1}{c|}{0.98} & 0.91 &
\multicolumn{1}{c|}{\cellcolor[HTML]{FFEAE9}IONIQ 5}   &
\multicolumn{1}{c|}{16.2} & \multicolumn{1}{c|}{9.6} & \multicolumn{1}{c|}{5.2} &
\multicolumn{1}{c|}{1.0}  & \multicolumn{1}{c|}{0.08} & 0.13 \\ \hline

\multicolumn{7}{|c|}{\cellcolor[HTML]{9AFF99}NLL Drop Out} &
\multicolumn{7}{c|}{\cellcolor[HTML]{9AFF99}Quantile Anchor} \\ \hline

\multicolumn{1}{|c|}{\cellcolor[HTML]{CBCEFB}Vehicle} &
\multicolumn{1}{c|}{\cellcolor[HTML]{DCFFDC}RMSE} &
\multicolumn{1}{c|}{\cellcolor[HTML]{DCFFDC}MAE} &
\multicolumn{1}{c|}{\cellcolor[HTML]{DCFFDC}AU}  &
\multicolumn{1}{c|}{\cellcolor[HTML]{DCFFDC}EU}  &
\multicolumn{1}{c|}{\cellcolor[HTML]{DCFFDC}$R^2$}  &
\cellcolor[HTML]{DCFFDC}EV &
\multicolumn{1}{c|}{\cellcolor[HTML]{CBCEFB}Vehicle} &
\multicolumn{1}{c|}{\cellcolor[HTML]{DCFFDC}RMSE} &
\multicolumn{1}{c|}{\cellcolor[HTML]{DCFFDC}MAE} &
\multicolumn{1}{c|}{\cellcolor[HTML]{DCFFDC}AU}  &
\multicolumn{1}{c|}{\cellcolor[HTML]{DCFFDC}EU}  &
\multicolumn{1}{c|}{\cellcolor[HTML]{DCFFDC}$R^2$}  &
\cellcolor[HTML]{DCFFDC}EV \\ \hline

\multicolumn{1}{|c|}{\cellcolor[HTML]{FFEAE9}Sedan\_1}  &
\multicolumn{1}{c|}{2.5} & \multicolumn{1}{c|}{1.7} & \multicolumn{1}{c|}{1.7} &
\multicolumn{1}{c|}{1.5} & \multicolumn{1}{c|}{0.93} & 0.94 &
\multicolumn{1}{c|}{\cellcolor[HTML]{FFEAE9}Sedan\_1}  &
\multicolumn{1}{c|}{3.9} & \multicolumn{1}{c|}{2.5} & \multicolumn{1}{c|}{1.5} &
\multicolumn{1}{c|}{10.2} & \multicolumn{1}{c|}{0.83} & 0.85 \\ \hline

\multicolumn{1}{|c|}{\cellcolor[HTML]{FFEAE9}Sedan\_2}  &
\multicolumn{1}{c|}{2.7} & \multicolumn{1}{c|}{2.1} & \multicolumn{1}{c|}{1.9} &
\multicolumn{1}{c|}{1.8} & \multicolumn{1}{c|}{0.90} & 0.90 &
\multicolumn{1}{c|}{\cellcolor[HTML]{FFEAE9}Sedan\_2}  &
\multicolumn{1}{c|}{2.8} & \multicolumn{1}{c|}{1.8} & \multicolumn{1}{c|}{5.5} &
\multicolumn{1}{c|}{11.1} & \multicolumn{1}{c|}{0.85} & 0.85 \\ \hline

\multicolumn{1}{|c|}{\cellcolor[HTML]{FFEAE9}Truck}     &
\multicolumn{1}{c|}{5.0} & \multicolumn{1}{c|}{3.2} & \multicolumn{1}{c|}{2.2} &
\multicolumn{1}{c|}{1.8} & \multicolumn{1}{c|}{0.89} & 0.90 &
\multicolumn{1}{c|}{\cellcolor[HTML]{FFEAE9}Truck}     &
\multicolumn{1}{c|}{5.7} & \multicolumn{1}{c|}{3.5} & \multicolumn{1}{c|}{1.8} &
\multicolumn{1}{c|}{16.0} & \multicolumn{1}{c|}{0.85} & 0.85 \\ \hline

\multicolumn{1}{|c|}{\cellcolor[HTML]{FFEAE9}Hatchback} &
\multicolumn{1}{c|}{2.7} & \multicolumn{1}{c|}{1.8} & \multicolumn{1}{c|}{1.3} &
\multicolumn{1}{c|}{1.4} & \multicolumn{1}{c|}{0.91} & 0.91 &
\multicolumn{1}{c|}{\cellcolor[HTML]{FFEAE9}Hatchback} &
\multicolumn{1}{c|}{3.2} & \multicolumn{1}{c|}{2.3} & \multicolumn{1}{c|}{0.90} &
\multicolumn{1}{c|}{8.7} & \multicolumn{1}{c|}{0.87} & 0.87 \\ \hline

\multicolumn{1}{|c|}{\cellcolor[HTML]{FFEAE9}IONIQ 5}   &
\multicolumn{1}{c|}{7.7} & \multicolumn{1}{c|}{3.3} & \multicolumn{1}{c|}{4.8} &
\multicolumn{1}{c|}{1.8} & \multicolumn{1}{c|}{0.78} & 0.79 &
\multicolumn{1}{c|}{\cellcolor[HTML]{FFEAE9}IONIQ 5}   &
\multicolumn{1}{c|}{8.6} & \multicolumn{1}{c|}{3.7} & \multicolumn{1}{c|}{3.3} &
\multicolumn{1}{c|}{28}  & \multicolumn{1}{c|}{0.73} & 0.75 \\ \hline
\end{tabular}
\label{tab:full_report_vehicles}
\caption{Comprehensive results across five vehicle classes—two sedans, one pickup truck, one hatchback, and a Hyundai IONIQ 5—comparing four uncertainty heads: \emph{t–NLL Anchor}, \emph{t–NLL Dropout}, \emph{Quantile Anchor}, and \emph{Quantile MC Dropout}.}
\end{table*}

\begin{table}[]
\centering
\begin{tabular}{|l|lllll|}
\hline
\rowcolor[HTML]{67FD9A} 
\cellcolor[HTML]{FFFFFF}                        & \multicolumn{5}{|c|}{\cellcolor[HTML]{9AFF99}Sedan}                                                                                                                                                                           \\ \cline{2-6} 
\rowcolor[HTML]{DCFFDC} 
\multirow{-2}{*}{\cellcolor[HTML]{FFFFFF}Model} & \multicolumn{1}{l|}{\cellcolor[HTML]{DCFFDC}Convergence} & \multicolumn{1}{l|}{\cellcolor[HTML]{DCFFDC}Low} & \multicolumn{1}{l|}{\cellcolor[HTML]{DCFFDC}High} & \multicolumn{1}{l|}{\cellcolor[HTML]{DCFFDC}Width} & StVar \\ \hline
\cellcolor[HTML]{ECF4FF}NLL Anchor              & \multicolumn{1}{l|}{0.898}                               & \multicolumn{1}{l|}{0.889}                       & \multicolumn{1}{l|}{0.907}                        & \multicolumn{1}{l|}{5.03}                          & 1.13  \\ \hline
\cellcolor[HTML]{ECF4FF}NLL Dropout             & \multicolumn{1}{l|}{0.911}                               & \multicolumn{1}{l|}{0.902}                       & \multicolumn{1}{l|}{0.919}                        & \multicolumn{1}{l|}{7.69}                          & 0.948 \\ \hline
\cellcolor[HTML]{ECF4FF}Quantile Anchor         & \multicolumn{1}{l|}{0.966}                               & \multicolumn{1}{l|}{0.960}                       & \multicolumn{1}{l|}{0.971}                        & \multicolumn{1}{l|}{32.32}                         & 0.36  \\ \hline
\cellcolor[HTML]{ECF4FF}Quantile Dropout        & \multicolumn{1}{l|}{1.000}                               & \multicolumn{1}{l|}{0.999}                       & \multicolumn{1}{l|}{1.000}                        & \multicolumn{1}{l|}{27.59}                         & 0.06  \\ \hline
Model                                           & \multicolumn{5}{|c|}{\cellcolor[HTML]{9AFF99}IONIQ 5}                                                                                                                                                                            \\ \hline
\cellcolor[HTML]{ECF4FF}NLL Anchor              & \multicolumn{1}{l|}{0.919}                               & \multicolumn{1}{l|}{0.885}                       & \multicolumn{1}{l|}{0.944}                        & \multicolumn{1}{l|}{12.59}                         & 1.13  \\ \hline
\cellcolor[HTML]{ECF4FF}NLL Dropout             & \multicolumn{1}{l|}{0.949}                               & \multicolumn{1}{l|}{0.920}                       & \multicolumn{1}{l|}{0.968}                        & \multicolumn{1}{l|}{18.48}                         & 1.45  \\ \hline
\cellcolor[HTML]{ECF4FF}Quantile Anchor         & \multicolumn{1}{l|}{0.991}                               & \multicolumn{1}{l|}{0.974}                       & \multicolumn{1}{l|}{0.996}                        & \multicolumn{1}{l|}{95.53}                         & 0.11  \\ \hline
\cellcolor[HTML]{ECF4FF}Quantile Dropout        & \multicolumn{1}{l|}{0.327}                               & \multicolumn{1}{l|}{0.279}                       & \multicolumn{1}{l|}{0.379}                        & \multicolumn{1}{l|}{17.89}                         & 9.15  \\ \hline
\end{tabular}
\caption{Confidence-interval analysis. ‘Low’ and ‘High’ are the lower and upper bounds of the binomial coverage band; ‘Width’ is the mean interval width; ‘StVar’ is the variance of standardized residuals.}
\label{tab:confidence}
\end{table}

\section{Discussion} \label{sec:discussion}
\subsection{Theoretical Discussion}
We presented an \emph{anchored LSTM} framework for EV power estimation that delivers both point predictions and calibrated uncertainty, jointly capturing \emph{epistemic} (model) and \emph{aleatoric} (data) uncertainty components. Beyond empirical performance gains, our contribution provides a clean, end-to-end probabilistic formulation with explicit priors and distribution-aware uncertainty quantification. Specifically:

\begin{itemize}
\item \textbf{Anchoring for LSTM networks:} We extended the Bayesian anchoring method to LSTM architectures. This approach yields a \emph{posterior-approximating ensemble} by sampling anchors $\mathbf{w}_0^{(m)}$ and retraining, establishing an explicit connection between ensemble diversity and epistemic uncertainty.

\item \textbf{Heavy-tailed likelihood with analytical intervals:} By employing the Student's-t distribution as a loss function, the predictive interval at confidence level $1-\alpha$ has an \emph{analytical form}: $I_{1-\alpha}(x)=\mu(x)\pm t_{1-\alpha/2,\nu}$. This formulation provides provable robustness to outliers compared to Gaussian negative log-likelihood while maintaining likelihood-based calibration.

\item \textbf{Explicit uncertainty separation:} The anchored ensemble quantifies epistemic uncertainty through between-member variance, while the learned scale parameter within the loss function captures aleatoric noise.
\end{itemize}

This framework yields a model that is (i) \emph{principled}—implementing MAP estimation with a proper heavy-tailed likelihood and exact 
t
t-distribution quantile intervals; (ii) \emph{diagnosable}—enabling calibration verification against known quantiles and binomial coverage scores; and (iii) \emph{efficient}—eliminating test-time sampling requirements inherent in MC dropout approaches.

\subsection{Practical Implications and Model Performance}
In this study, we demonstrated that the NLL-anchored model effectively captures variability in battery power estimation. To our knowledge, this represents the first work to separately quantify both aleatoric and epistemic uncertainties while developing a mathematically rigorous probabilistic approach for epistemic uncertainty quantification.
Our key findings indicate that the proposed model provides well-calibrated confidence intervals. As expected, real-world highway data exhibited slightly increased RMSE and MAE values compared to chassis dynamometer experiments; however, the results remain strong. Through binomial proportion tests, we have verified that the uncertainty bands are well-calibrated, indicating high confidence in the model's predictions.

The existing literature on EV battery energy estimation predominantly employs quantile loss for capturing aleatoric noise \cite{chen2021data, de2017data}. We used quantile regression as a benchmark against our proposed Student's-t distribution loss function. Notably, for the noisier IONIQ 5 highway data, quantile regression performed substantially worse. The model became overly conservative, producing excessively wide confidence bands that were nearly meaningless, with standardized variance values around 0.1. In contrast, the Student's-t loss function demonstrated significantly better performance across all datasets. Even in controlled chassis dynamometer experiments, quantile regression yielded inferior results.
We attribute this performance difference to several factors: empirically, quantile-based variants produced systematically wider intervals despite comparable RMSE/MAE. This behavior is consistent with (1) a more challenging optimization landscape compared to MSE-like objectives, (2) multi-quantile training ($\tau\in{0.1,0.5,0.9}$) creating a complex multi-task learning problem, and (3) the nonparametric nature of quantile regression, which lacks distributional priors and tends toward conservative coverage at the expense of precision.

Compared to MC dropout, the anchored LSTM achieves similar accuracy and calibration while offering significant practical advantages: inference does not require stochastic sampling during test time. This makes the anchored approach particularly suitable for real-time energy management applications with strict latency constraints. Furthermore, the ensemble is trained once and evaluated with a single forward pass per member (or aggregated via weight averaging), enabling straightforward scaling to larger datasets and multi-vehicle scenarios without incurring test-time Monte Carlo overhead.

Overall, our results demonstrate that the anchored LSTM with a Student's-$t$ output head represents a practical and reliable alternative to both MC dropout and quantile methods for uncertainty-aware EV power prediction, combining strong accuracy, competitive calibration, and deployment-ready efficiency.

While this study serves as a proof-of-concept for learning calibrated epistemic and aleatoric uncertainties using both chassis dynamometer and highway driving data, a limitation remains in the dataset size for developing a fully generalized model. Expanding to larger, more diverse datasets represents an important direction for future work to enhance model generalization across varying driving conditions and vehicle types.

\section{Conclusion}\label{sec:conclusion}

In this study, we addressed the critical challenge of range anxiety in electric vehicles by developing an accurate power consumption estimation framework based on vehicle kinematics. Reliable power prediction is essential for optimizing battery charging schedules, developing energy-efficient driving strategies, and enabling effective real-time energy management and route planning. For these applications to be trustworthy, power consumption predictions must be accompanied by well-calibrated confidence intervals.

To this end, we implemented an anchored Bayesian ensemble approach. While originally developed for standard feedforward neural networks, we demonstrated that this methodology extends naturally to LSTM architectures, providing a principled framework for uncertainty quantification in sequential data modeling.

Our model achieved excellent performance across all five vehicle fleets (combining chassis dynamometer and highway data), with an average RMSE of \(3.36 \pm 1.1\), MAE of \(2.21 \pm 0.89\), \(R^2\) of \(0.93 \pm 0.02\), and explained variance of \(0.93 \pm 0.02\). More importantly, we demonstrated that the model provides meaningful, well-calibrated uncertainty bands that reliably capture both epistemic and aleatoric uncertainties.

In comprehensive benchmarking against alternative approaches, our method performed comparably to or better than existing techniques. While achieving similar accuracy to MC dropout, our anchored ensemble offers significant practical advantages—particularly for real-time applications—by eliminating the need for stochastic sampling during inference. This makes our approach more suitable for deployment in latency-sensitive energy management systems.

The framework presented in this work provides a solid foundation for uncertainty-aware EV power prediction, balancing theoretical rigor with practical deployability to help alleviate range anxiety through reliable energy forecasting.

\section*{ACKNOWLEDGMENT}
This work was co-funded by Transport Canada’s ecoTECHNOLOGY for Vehicles program and the National Research Council Canada’s Clean and Energy Efficient Transportation program. The views and opinions of the authors expressed herein do not necessarily state or reflect those of Transport Canada.

\bibliographystyle{IEEEtran}
\bibliography{references.bib}

\newpage

\section*{Appendix} \label{sec:appendic}

We begin by specifying the LSTM architecture in Section~\ref{sec:LSTM}. Subsequently, Section~\ref{sec:prop} generalizes the anchored MAP framework to this model and outlines a sketch of the proof.
\subsection*{LSTM} \label{sec:LSTM}
For sequential data, LSTM networks are effective due to their ability to retain temporal dependencies, which is useful for energy estimation from time-series inputs (e.g., velocity, torque, acceleration). An LSTM cell comprises input, forget, and output gates that regulate information flow. At time $t$, given input $x_t$ and previous hidden state $h_{t-1}$,
\begin{align*}
\label{LSTM_gates}
i_t &= \sigma\!\big(W_{xi} x_t + b_{i} + W_{hi} h_{t-1} + b_{hi}\big), \\
f_t &= \sigma\!\big(W_{xf} x_t + b_{f} + W_{hf} h_{t-1} + b_{hf}\big), \\
o_t &= \sigma\!\big(W_{xo} x_t + b_{o} + W_{ho} h_{t-1} + b_{ho}\big), \\
\tilde c_t &= \tanh\!\big(W_{xc} x_t + b_{c} + W_{hc} h_{t-1} + b_{hc}\big), \\
c_t &= f_t \odot c_{t-1} + i_t \odot \tilde c_t, \\
h_t &= o_t \odot \tanh(c_t),
\end{align*}
where $\sigma(\cdot)$ is the logistic function; $W_{\{\cdot\}}$ and $b_{\{\cdot\}}$ are gate-specific weights and biases; and $\odot$ denotes elementwise multiplication. In our EV power estimation context, $x_t$ represents vehicle operational parameters and the target $y$ is instantaneous power.

\noindent Building on the partition $W=\bigoplus_{g\in\mathcal{G}}W_g$ defined above, we
place independent Gaussian priors (or anchors) on each block $W_g$ and optimize MAP.

\subsection*{Extension of Anchor model into LSTM architecture} \label{sec:prop}
\paragraph*{\textbf{Proposition (MAP with gate-wise Gaussian priors)}}
Let $\mathcal{D}=\{(x_n,y_n)\}_{n=1}^N$ and let $\hat y(\cdot;W)$ denote the LSTM predictor with
\[
W \;=\; \bigoplus_{g\in\mathcal{G}} W_g, \qquad 
\mathcal{G}=\{\text{input }i,\ \text{forget }f,\ \text{output }o,\ \text{candidate }c,\ \text{head}\}.
\]
Assume the likelihood factorizes over samples, $p(\mathcal{D}\mid W)=\prod_{n=1}^N p(y_n\mid x_n,W)$,
and the prior factorizes over blocks,
\[
p(W) \;=\; \prod_{g\in\mathcal{G}} \mathcal{N}\!\big(W_g;\,W_g^{(0)},\,\sigma_p^2 I\big).
\]
Define the empirical data term
\[
\mathcal{L}_{\mathrm{data}}(W)=\frac{1}{N}\sum_{n=1}^N \ell\!\big(\hat y(x_n;W),\,y_n\big),
\quad \text{with } \ell(\hat y,y)=-\log p(y\mid \hat y).
\]
Then any MAP estimator minimizes
\[
\boxed{\;
\mathcal{J}(W)=\mathcal{L}_{\mathrm{data}}(W)+\frac{1}{2\sigma_p^{2}}\sum_{g\in\mathcal{G}} \big\|W_g - W_g^{(0)}\big\|_2^2
\;}
\]
i.e., the loss is a function of all gates plus a Gaussian quadratic penalty applied separately to each block.

\paragraph*{\textbf{Proof.}}
The average negative log-posterior is
\[
-\tfrac{1}{N}\log p(W\mid\mathcal{D})
=\tfrac{1}{N}\sum_{n=1}^N \big[-\log p(y_n\mid x_n,W)\big] - \tfrac{1}{N}\log p(W) + C,
\]
and $\ell(\hat y(x_n;W),y_n)=-\log p(y_n\mid x_n,W)$ gives the data term.
Using $-\log \mathcal{N}(w;\mu,\sigma^2 I)=\tfrac{1}{2\sigma^2}\|w-\mu\|_2^2+C$ and prior independence,
\[
-\log p(W)=\sum_{g\in\mathcal{G}} \tfrac{1}{2\sigma_p^{2}} \|W_g - W_g^{(0)}\|_2^2 + C',
\]
so minimizing $-\log p(W\mid\mathcal{D})$ is equivalent (up to constants) to minimizing $\mathcal{J}(W)$. \qed

\end{document}